\documentclass[10pt,twocolumn,letterpaper]{article}
\usepackage[accsupp]{axessibility}
\usepackage{cvpr}              

\definecolor{cvprblue}{rgb}{0.21,0.49,0.74}
\usepackage[pagebackref,breaklinks,colorlinks,allcolors=cvprblue]{hyperref}
\usepackage[ruled,vlined]{algorithm2e}  
\usepackage[utf8]{inputenc}   
\usepackage[T1]{fontenc}       
\usepackage{amsmath}           
\usepackage{algpseudocode}
\usepackage{multirow}
\usepackage{pifont}
\usepackage{colortbl}
\usepackage{graphicx}
\usepackage{float}
\usepackage{caption}
\usepackage{subcaption}
\def\paperID{64} 
\def\confName{CVPR}
\def\confYear{2025}
\hypersetup{
    colorlinks=true, 
    urlcolor=magenta,   
}
\title{Neighbor-Based Feature and Index Enhancement for Person Re-Identification}

\author{
Chao Yuan$^{1,*}$, Tianyi Zhang$^{2,*}$, Guanglin Niu$^{2,\dagger}$ \\
\textsuperscript{1} School of Computer Science and Engineering, Beihang University \quad \\
\textsuperscript{2} School of Artificial Intelligence, Beihang University  \\
{\tt\small yuanc3666@gmail.com}, {\tt\small \{zy2442222,beihangngl\}@buaa.edu.cn} \\
}

\begin{document}

\maketitle
\renewcommand{\thefootnote}{}
\footnotetext{* Equal contribution. $\dagger$ Corresponding author.}
\footnotetext{Codes: \textit{\url{https://github.com/yuanc3/DMON-ARO}}}



\begin{abstract}
Person re-identification (Re-ID) aims to match the same pedestrian in a large gallery with different cameras and views. Enhancing the robustness of the extracted feature representations is a main challenge in Re-ID. Existing methods usually improve feature representation by improving model architecture, but most methods ignore the potential contextual information, which limits the effectiveness of feature representation and retrieval performance. Neighborhood information, especially the potential information of multi-order neighborhoods, can effectively enrich feature expression and improve retrieval accuracy, but this has not been fully explored in existing research. Therefore, we propose a novel model \textbf{DMON-ARO} that leverages latent neighborhood information to enhance both feature representation and index performance. Our approach is built on two complementary modules: Dynamic Multi-Order Neighbor Modeling (DMON) and Asymmetric Relationship Optimization (ARO). The DMON module dynamically aggregates multi-order neighbor relationships, allowing it to capture richer contextual information and enhance feature representation through adaptive neighborhood modeling. Meanwhile, ARO refines the distance matrix by optimizing query-to-gallery relationships, improving the index accuracy. Extensive experiments on three benchmark datasets demonstrate that our approach achieves performance improvements against baseline models, which illustrate the effectiveness of our model. Specifically, our model demonstrates improvements in Rank-1 accuracy and mAP. Moreover, this method can also be directly extended to other re-identification tasks.
\end{abstract}


\section{Introduction}
\label{sec:intro}
\begin{figure}[htbp]
\centering
 \includegraphics[width=0.5\textwidth]{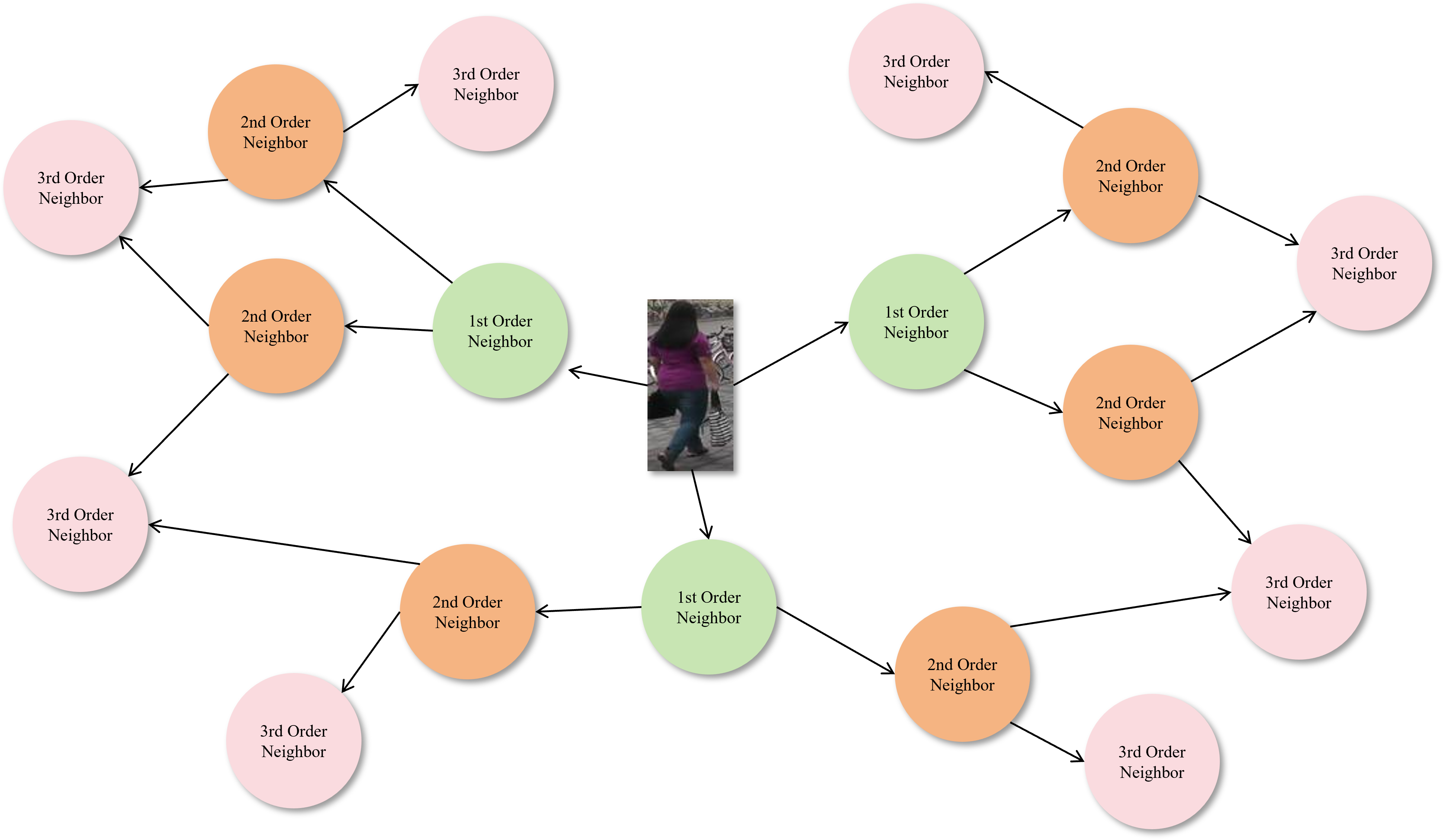}
\caption{Multi-Order Neighbor Structure for Feature Enhancement in Person Re-Identification.}
\label{fig:k}
\end{figure}
\begin{figure*}[htbp]
\centering
 \includegraphics[width=1\textwidth]{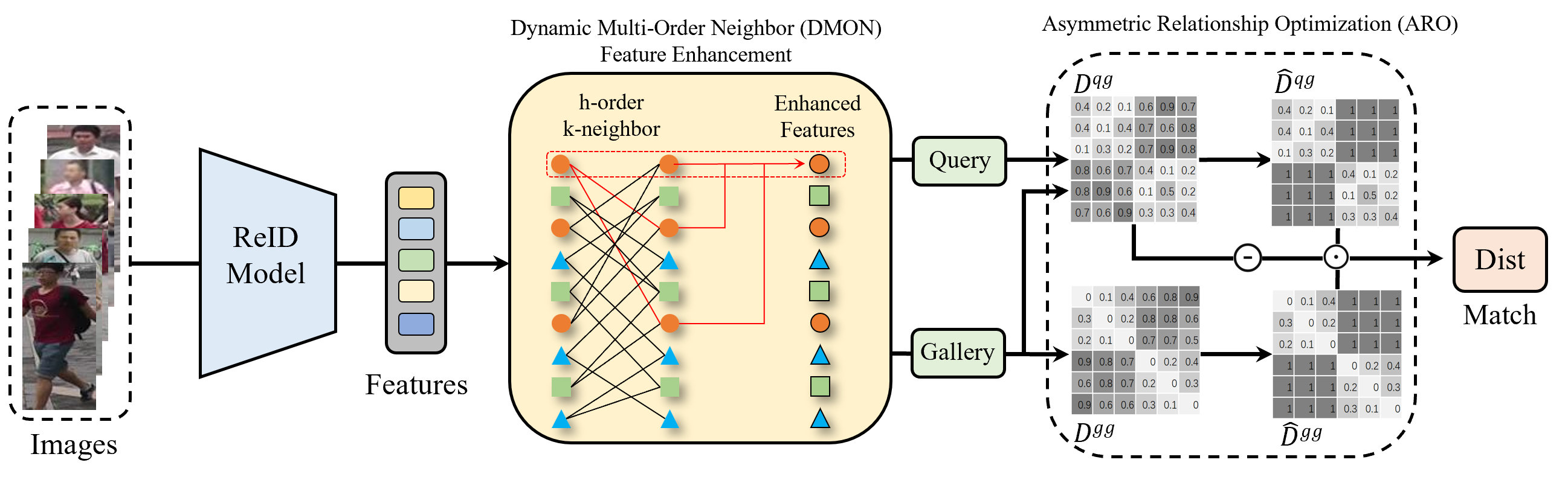}
\caption{Framework of our proposed methods. Dynamic Multi-Order Neighbor (DMON) searches the potential neighbors to enhance feature representation. Asymmetric Relationship Optimization (ARO) using query-gallery relations to optimize the distance matrix for better matching.}
\label{fig:main}
\end{figure*}
Person Re-Identification (ReID) aims to accurately identify and match the same pedestrian across different cameras~\cite{8099872}, which has garnered increasing attention from both academia and industry. Whereas, existing approaches still face significant challenges in practical scenarios due to factors such as viewpoint variations, occlusions, lighting differences, and background complexity.

Most existing Re-ID models~\cite{wang2018learning} focus on improving recognition performance by refining model architectures, such as modeling global and local features to enhance feature representation. However, these models often ignore potential contextual information, particularly that derived from neighbor samples, leading to suboptimal feature expressiveness and retrieval performance. Leveraging information from neighbor samples to enhance both feature representation and retrieval performance is a significant yet underexplored direction.

To address this issue, this paper proposes a Re-ID strategy \textbf{DMON-ARO} which consists of two main modules, namely Dynamic Multi-Order Neighbor (DMON) Feature Enhancement and Asymmetric Relation Optimization (ARO). The goal of our model is to exploit neighbor sample information to improve feature expressiveness and retrieval robustness.

Specifically, DMON constructs first-order, second-order, and third-order neighbor sets by recursively expanding neighbor relationships. A dynamic adaptive weighting mechanism assigns appropriate weights to different-order neighbors, enabling the model to capture contextual information more effectively and thereby enhancing feature representation. This multi-order neighbor construction not only facilitates effective smoothing of the feature space, ensuring that intra-class samples become more compact and inter-class samples remain distinguishable, but also enhances the model's ability to handle sparse or imbalanced data. By leveraging higher-order neighbors, DMON can compensate for the lack of information in sparse regions of the sample space, improving robustness and generalization in challenging scenarios.

Furthermore, ARO optimizes the feature representation by leveraging neighbor information. Unlike traditional fixed-distance metrics such as Euclidean distance or cosine similarity~\cite{schroff2015facenet}, ARO introduces neighbor relationship modeling between query and gallery features to adaptively optimize similarity measurement. Thus, ARO could enhance the discriminative capability of the features, thereby improving the retrieval performance in complex scenarios. 

Our contributions can be summarized as follows:

\begin{itemize}
\item This is the first effort to take advantage of multi-order neighbors in Re-ID. Our model could effectively leverage multi-order neighbor information to enhance feature representation with an adaptive weighting mechanism.

\item We propose an asymmetric similarity relationship optimization strategy that utilizes neighbor information to further enhance the representation of both query and gallery features, improving retrieval performance.

\item Extensive experiments on three standard datasets show significant performance improvements of our model compared to some state-of-the-art baseline models, illustrating the effectiveness of the proposed modules.
\end{itemize}









\section{Related works}

\subsection{Person Re-Identification}

Person Re-Identification (Re-ID) is a pivotal research area within the domain of computer vision, aiming to identify and retrieve images of the same pedestrian across different camera views from a database of candidate images. Some representative models are declared in the following. IDE~\cite{7410490} formulates the Re-ID task as a multi-class classification problem. Each unique identity is treated as a separate class, and a classification network, such as ResNet, is employed to extract global features. During testing, identity retrieval is performed through nearest neighbor matching based on these features. PCB~\cite{sun2018beyond} partitions the image into distinct regions, allowing the model to concentrate on various body parts of an individual. This approach effectively reduces the impact of occlusions and pose variations on the re-identification process. TransReID~\cite{he2021transreid} introduces a vision Transformer-based feature modeling technique for person Re-ID~\cite{dosovitskiy2020image}. By leveraging the self-attention mechanism~\cite{vaswani2017attention} of Transformers, TransReID divides input images into patches and processes them individually. This strategy enables the model to capture both long-range contextual information and fine-grained local features.
Recent work proposes a Training-Free Feature Centralization framework (Pose2ID)~\cite{yuan2025poses} that can be directly applied to different ReID tasks and models, even an ImageNet pre-trained model without ReID training.
Some advances in person re-identification (Re-ID) focus on handling clothing changes. IFD~\cite{xu2025ifd} adopts a two-stream architecture, where the attention stream processes the clothing mask image and generates identity attention weights to effectively transfer spatial knowledge to the main stream and highlight the regions containing rich identity information. A scalable pre-training-fine-tuning framework improves the ReID model after pre-training on CCUP~\cite{zhao2024ccup} and fine-tuning on benchmark datasets.

\subsection{Neighborhood Modeling}

To further enhance the adaptability of models in complex scenarios, neighbor information has gained significant traction in Re-ID tasks in recent years. This information effectively models the relationships between samples in the feature space, thereby improving the model's discriminative ability and robustness. For instance, KNN~\cite{8099872} enhances re-identification robustness by reordering the candidate set based on the intersection and distances of k-nearest neighbors. GCN~\cite{shen2018deep} propagates feature information through neighbor relationships within a graph, capturing more intricate global contextual relationships. CFA~\cite{Sarfraz2017APE} leverages the contextual information of neighboring samples to enhance the features of target samples. By averaging or weighted aggregating neighbor features, the representation of target samples becomes more comprehensive. MBNC~\cite{9157237} introduces a memory module that stores neighbor features during the training phase, dynamically adjusting the feature representation of query samples by leveraging these neighbor features. Memory updates are further refined using neighbor relationships, which enhance feature robustness through shared neighbor connections. In contrast to these methods, our proposed approach constructs multi-order neighborhood relationships to more effectively exploit global contextual information.



\section{Methods}
As illustrated in Fig.\ref{fig:main}, the proposed framework consists of two main modules: the Dynamic Multi-Order Neighbor Feature Enhancement (DMON) Module and the Asymmetric Relation Optimization (ARO) Module. The Asymmetric Relation Optimization Module further models asymmetric matching relationships between query and candidate features, thereby enhancing the accuracy of feature matching.

Specifically, we utilize the pre-trained ViT-TransReID model as the baseline. First, query and gallery images are input into the model to compute the initial features. Then, the DMON mechanism is applied to get enhanced feature representations for both query features and gallery features. Finally, the ARO Module computes the distance matrix, and the optimized distance matrix is used to output the final feature matching results.

\subsection{Dynamic Multi-Order Neighbor Feature Enhancement}

The goal of the DMON module is to improve the robustness of feature representation by capturing global contextual information from multi-order neighbor relationships. Thus, this module is based on the construction of multi-order neighbors with a dynamic weight allocation strategy. Firstly, we calculate the Euclidean distance matrix \( D \in \mathbb{R}^{N \times N} \) between the samples as:
\begin{equation}
D_{ij} = \| F_i - F_j \|_2
\end{equation}
where the input feature is denoted as \( F \in \mathbb{R}^{N \times D} \). \( N \) is the number of samples and \( D \) is the dimension of feature.

Then, we construct multi-order neighbor relationships iteratively, starting from the first-order neighbors and gradually expanding to second-order and third-order neighbors. Each order of neighbors is the union of the nearest neighbors of the previous order, thus capturing higher-order global feature associations. Given the distance matrix \( D \), we construct the first-order, second-order, and third-order neighbor relationships.

\textbf{First-order neighbors.}
For each sample \( x \), we select the top \( k_1 \) nearest neighbors as its first-order neighbors:
\begin{equation}
N^{(1)}(x) = \{ y \mid y \in \text{Top-}k_1(D(x, Y)) \}
\end{equation}

\textbf{Higher-order neighbor expansion.}
The second-order and higher order neighbors \(N^{(h)}\) are further expanded from the former order neighbors \(N^{(h-1)}\). For each first-order neighbor \( y \in N^{(1)}(x) \), we select samples from the top \( k_1 \) nearest neighbors of \( y \) as the second-order neighbors of \( x \):
\begin{equation}
N^{(h)}(x) = \bigcup_{y \in N^{(h-1)}(x)} N^{(1)}(y)
\end{equation}

We then design a dynamic weight allocation mechanism based on the Gaussian kernel for neighbors at different levels. The weights are dynamically adjusted according to the distance between neighboring samples, ensuring that the contribution of neighbors to feature enhancement aligns with the sample distribution characteristics.

Based on the Gaussian kernel function, we assign weights to each level of neighbors:
\begin{equation}
W_{xy}^{(h)} = \exp\left(-\frac{1}{2\sigma_h^2} D_{xy}^2 \right) \cdot M_{xy}
\end{equation}
where \( \sigma_h = \sigma \cdot (1.5)^h \) is the kernel bandwidth that increases with the level, controlling the range of the weight distribution.

We compute the latent features of each-order neighbors by summing them with decay coefficients, and linearly fuse them with the original features to generate the final feature representation.

The enhanced feature of each-order neighbor is:
\begin{equation}
F^{(h)} = W^{(h)} \cdot F
\end{equation}
where \( F^{(h)} \) denotes the feature representation of the \( h \)-th order neighbor. The features of different-order neighbors are combined with decay coefficients:
\begin{equation}
F_{\text{latent}} = \sum_{h=1}^{H} \alpha_h F^{(h)}
\end{equation}
where \( \alpha_h \in [1, 0.5, 0.25,...]\) represents the decay coefficient of the \( h \)-th order neighbor, ensuring that the contribution of higher-order neighbors gradually diminishes.

The enhanced feature representation is obtained by linearly weighting the original and latent features:
\begin{equation}
F = \| \gamma \cdot F + (1 - \gamma) \cdot F_{\text{latent}} \|_2
\end{equation}

\subsection{Asymmetric Relationship Optimization (ARO)}

The goal of the Asymmetric Relationship Optimization module is to explicitly model the asymmetric distribution differences between features by separately optimizing the matching relationships of query and gallery features, thereby improving the matching accuracy for ReID. This asymmetric optimization addresses the inherent modal differences between the query and gallery features, which exhibit distinct distributions and require independent modeling.

First, the initial distance matrix \( D^{qg} \) between the query feature \( F^q \) and the gallery feature \( F^g \), as well as the self-similarity matrix \( D^{gg} \) within the gallery features, are computed. The formula
\begin{equation}
D^{qg}_{i,j} = \| F^q_i - F^g_j \|_2^2
\end{equation}
\begin{equation}
D^{gg}_{i,j}= \| F^g_i - F^g_j \|_2^2
\end{equation}
where \( \| \cdot \|_2 \) denotes the L2 norm. These matrices serve as the foundation for subsequent optimization. By separating \( D^{qg} \) and \( D^{gg} \), the module explicitly captures the asymmetric relationship between the cross-modal query-gallery pairs and the intra-modal gallery-gallery pairs.

To refine the initial distance matrices, the module focuses on local neighborhood relationships, where the nearest neighbors are identified to model pairwise dependencies.

\begin{algorithm}[h]

\caption{Asymmetric Relationship Optimization}
\label{alg:neighborhood_optimization}
\KwIn{Distance matrices $D^{qg}, D^{gg}$; neighborhood sizes $k_2$}
\KwOut{Optimized distance matrices $\hat{D}^{qg}$, $\hat{D}^{gg}$}

\textbf{Step 1: Neighborhood Selection}\;

\For{each query $i$ in $D^{qg}$}{
    Get top-$k_2$ nearest neighbors: \\$\text{rank}_{qg}[i] \gets \text{topk}(D^{qg}_{i}, k_2)$\;
}
\For{each gallery $i$ in $D^{gg}$}{
    Get top-$k_2$ nearest neighbors:\\ $\text{rank}_{gg}[i] \gets \text{topk}(D^{gg}_{i}, k_2)$\;
}

\textbf{Step 2: Distance Filtering}\;

\For{each query $i$ in $D^{qg}$}{
    $D^{qg}_{i,j} \gets 1$ if $j \notin \text{rank}_{qg}[i]$\;
}
\For{each gallery $i$ in $D^{gg}$}{
    $D^{gg}_{i,j} \gets 1$ if $j \notin \text{rank}_{gg}[i]$\;
}

\Return $\hat{D}^{qg}, \hat{D}^{gg}$\;
\end{algorithm}

This process reinforces the asymmetric optimization, as \( D^{qg} \) models cross-modal relationships while \( D^{gg} \) focuses on intra-modal relationships. Their independent handling avoids assuming symmetric distributions and better captures their unique characteristics.

Using the selected neighbors, the distances in \( D^{qg} \) and \( D^{gg} \) are filtered. For example, distances in \( D^{qg} \) exceeding the \( k_2 \)-th smallest distance are set to zero:
\begin{equation}
\hat{D}^{qg}_{i,j} = 
\begin{cases} 
D^{qg}_{i,j}, & \text{if } j \in \text{rank}_{qg}[i,:] \\ 
1, & \text{otherwise.} 
\end{cases}
\end{equation}

After neighborhood-based processing, the module constructs an asymmetric similarity matrix \( S \) to explicitly capture directional dependencies between query and gallery features:
\begin{equation}
A = \|\hat{D}^{qg}\|_2 \cdot \|\hat{D}^{gg}\|_2^T
\end{equation}
where \( \cdot \) denotes the matrix multiplication, and \( \| \cdot \|_2 \) denotes the L2 norm. \( A \) reflects the unidirectional influence of gallery self-similarities on query-gallery relationships, emphasizing the asymmetric dependency.

The original query-gallery distance matrix \( D^{qg} \) is adjusted using the computed similarity matrix \( A \), resulting in the final optimized distance matrix \( D_{\text{final}} \):
\begin{equation}
D_{\text{match}} = D^{qg} - A
\end{equation}

This final adjustment further highlights the asymmetric optimization, where the query-to-gallery distances are refined without enforcing a reciprocal influence from gallery-to-query.

\section{Experiments}
\subsection{Datasets and Evaluation Protocol}\label{AA}


\begin{table*}[htbp]
\caption{Comparison results (\%) with state-of-the-art baseline models. The best results are \textbf{bold}.}
\begin{center}
\renewcommand{\arraystretch}{1.4}
\renewcommand\tabcolsep{12pt}

\begin{tabular}{cc|cc|cc|cc}
\toprule
\multirow{2}{*}{Model} & \multirow{2}{*}{Venue} & \multicolumn{2}{c|}{Market-1501} & \multicolumn{2}{c|}{DukeMTMC} & \multicolumn{2}{c}{MSMT17} \\
                        &                        & mAP     & Rank-1     & mAP       & Rank-1     & mAP  & Rank-1   \\
\midrule
CAL~\cite{rao2021counterfactual}                     & ICCV 21                & 89.5         & 95.5            & 80.5           & 90               & 64         & 84.2          \\
ALDER*~\cite{zhang2021seeing}                  & TIP  21b               & 88.9         & 95.6            & 78.9           & 89.9             & 59.1       & 82.5          \\
CBDB-Net*~\cite{tan2021incomplete}               & TCSVT 21               & 85           & 94.4            & 74.3           & 87.7             & -          & -             \\
PCB~\cite{sun2018beyond}                     & ECCV 18                & 77.4         & 92.3            & 81.8           & 66.1             & -          & -             \\
MGN~\cite{wang2018learning}                     & MM   18                & 86.9         & 95.7            & \textbf{88.7}           & 78.4             & -          & -             \\
DRL-Net~\cite{9674853}                 & TMM  22                & 86.9         & 94.7            & 76.6           & 88.1             & 55.3       & 78.4          \\
LTReID*~\cite{wang2022ltreid}                 & TMM  22                & 89           & \textbf{95.9}            & 80.4           & 90.5             & 58.6       & 81            \\
MSINet~\cite{gu2023msinet}                  & CVPR 23                & 89.6         & 95.3            & -              & -                & 59.6       & 81            \\
KPR~\cite{Somers_2023}                     & ECCV 24                & 89.6         & \textbf{95.9}            & -              & -                & -          & -             \\
\midrule
TransReID~\cite{he2021transreid}               & ICCV 21                & 89.5         & 95.2            & 82.1           & 91.1             & 69.4       & 86.2          \\

TransReID + Ours         & -                      & \textbf{92.8}         & 95.4            & 87.7           & \textbf{91.5}             & \textbf{74.6}       & \textbf{86.4}          \\ 
\bottomrule
\end{tabular}
\label{tab:main_result}
\end{center}
\end{table*}

In this paper, we conduct experiments to verify the performance of our proposed model on three widely used pedestrian re-identification datasets: Market-1501~\cite{7410490}, MSMT17~\cite{MSMT17}, and DukeMTMC~\cite{Duke}. Market-1501 is one of the largest real-world benchmark datasets in the field of pedestrian re-identification with each individual appearing on a maximum of two cameras. MSMT17 contains the images captured from 15 different cameras with multiple perspectives and viewing angles. DukeMTMC is another popular pedestrian re-identification dataset sourced from the DukeMTMC camera network. 


To evaluate our approach, we use Rank-1 accuracy and mean Average Precision (mAP).

\subsection{Baseline Models}

We select several state-of-the-art baseline models for comparison, including CAL~\cite{rao2021counterfactual}, MSINet~\cite{gu2023msinet}, PCB~\cite{sun2018beyond}, MGN~\cite{wang2018learning}, DRL-Net~\cite{9674853}, LTReID~\cite{wang2022ltreid}, ALDER~\cite{zhang2021seeing}, CBDB-Net~\cite{tan2021incomplete}, KPR~\cite{Somers_2023} and TransReID~\cite{he2021transreid}. Particularly, TransReID is one of the most popular baseline approach, which is exploited as the basic model for training to extract features of each image. The experimental results of these baseline models are retrieved directly from the previous papers for fair comparison.
\begin{figure*}[htbp]
\centering
 \includegraphics[width=0.9\textwidth]{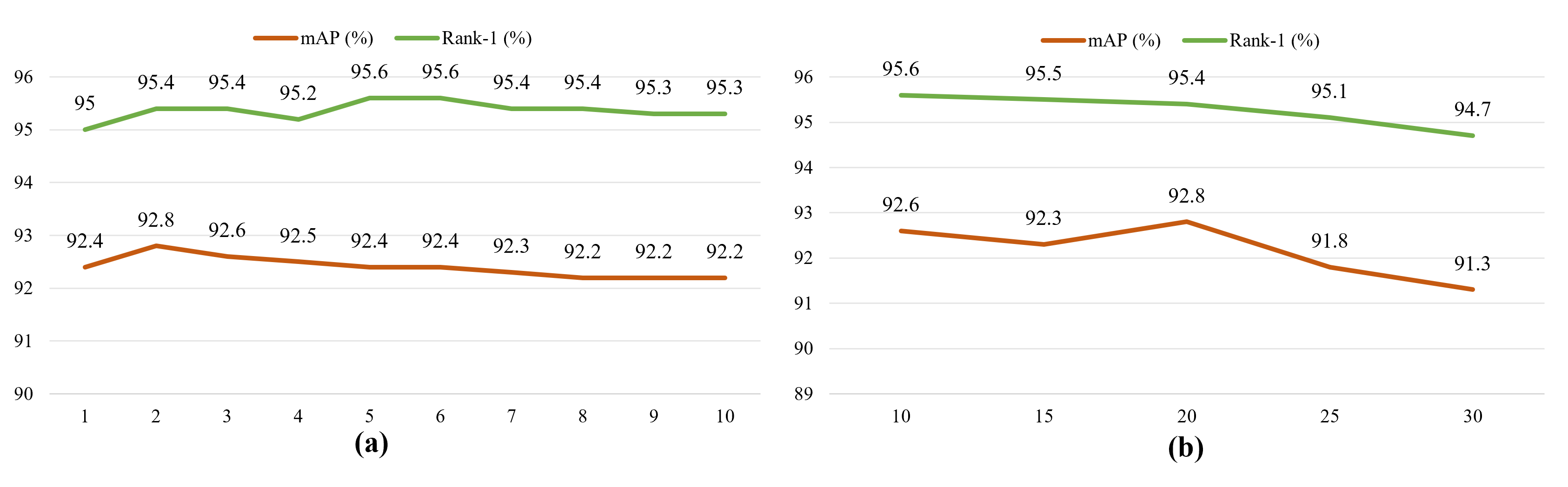}
\caption{(a) Impact of DMON Neighbor Parameter $k_1$ (left). (b) Impact of ARO Neighbor Parameter $k_2$ (right) on Market-1501 dataset.}
\label{fig:k}
\end{figure*}

\subsection{Implementation Details}
In this work, we build upon the ViT-TransReID model as our performance baseline. The input images are uniformly resized to a standard dimension of 256×128 pixels to maintain consistency across our experiments. We leverage the official pre-trained model weights for inferencing purposes. Our experimental framework is executed on an NVIDIA RTX 3090 GPU, ensuring robust computational support. The hyperparameter $\gamma$ is consistently set to 0.75 across all test scenarios. For the Market-1501 dataset, we set the values of $k_1$ and $k_2$ as 2 and 20, respectively. In contrast, for the DukeMTMC dataset, these values are adjusted to 5 and 20. Given the extensive size of the test images in the MSMT17 dataset, we have opted to partition the gallery set into batches of 10,000 images for processing by the DMON module. Correspondingly, the values of $k_1$ and $k_2$ for MSMT17 are set to 5 and 2, respectively.

\subsection{Comparison with State-of-the-Art Methods}
 We present a comparative analysis of our proposed Dynamic Multi-order Neighbors with Asymmetric Relation Optimization (DMON-ARO) method against current state-of-the-art techniques on the Market-1501, MSMT17, and DukeMTMC datasets. On Market-1501, our model achieves a Rank-1 accuracy of \textbf{95.4\%} and an mAP of \textbf{92.8\%}, showing a marginal but notable increase in Rank-1 accuracy by 0.2\% over TransReID (95.2\%) and by 0.1\% over MSINet (95.3\%), and a more substantial increase in mAP by 3.3\% over TransReID (89.5\%) and by 3.2\% over MSINet (89.6\%). These improvements indicate our model's enhanced robustness in handling variations in camera viewpoints, occlusions, and lighting conditions. On DukeMTMC, DMON-ARO achieves a Rank-1 accuracy of \textbf{91.5\%} and an mAP of \textbf{87.7\%}, outperforming the CAL method by 1.5\% in Rank-1 accuracy and by 7.2\% in mAP, and surpassing TransReID and CLIP-ReID by 5.6\% and 4.6\% in mAP, respectively. On the MSMT17 dataset, DMON-ARO achieves a Rank-1 accuracy of \textbf{86.4\%} and an mAP of \textbf{74.6\%}, which represents a 5.4\% increase in Rank-1 accuracy over MSINet (81\%) and a 2.2\% increase over CAL (84.2\%), and a 15\% increase in mAP over MSINet (59.6\%) and a 10.6\% increase over CAL (64\%). 
 
 These results illustrate the effectiveness of DMON-ARO in enhancing re-identification accuracy and precision across diverse and challenging datasets, highlighting its potential as a robust solution for person re-identification tasks.


\subsection{Parameter Analysis}

To assess the influence of varying parameter configurations on our proposed model, we conduct a comprehensive experimental analysis of the pivotal parameters on the dataset Market-1501. The neighbor parameter $k_1$ within our Dynamic Multi-order Neighbors (DMON) framework is of paramount importance, as it dictates the quantity of neighbors considered. As shown in Figure~\ref{fig:k}a, it is evident that when $k_1$ is minimal (e.g., $k_1=1$), the expansion of higher-order neighbors is inadequate, which detrimentally impacts the efficacy of feature enhancement. Optimal performance is achieved at $k_1=2$, with Rank-1 accuracy peaking at \textbf{95.4\%} and mAP at \textbf{92.6\%},where neighbor information is sufficiently expanded with minimal noise interference. As $k_1$ increases, an incremental accumulation of noise becomes evident, leading to a degradation of performance.

\begin{figure}[htbp]
\centering
 \includegraphics[width=0.45\textwidth]{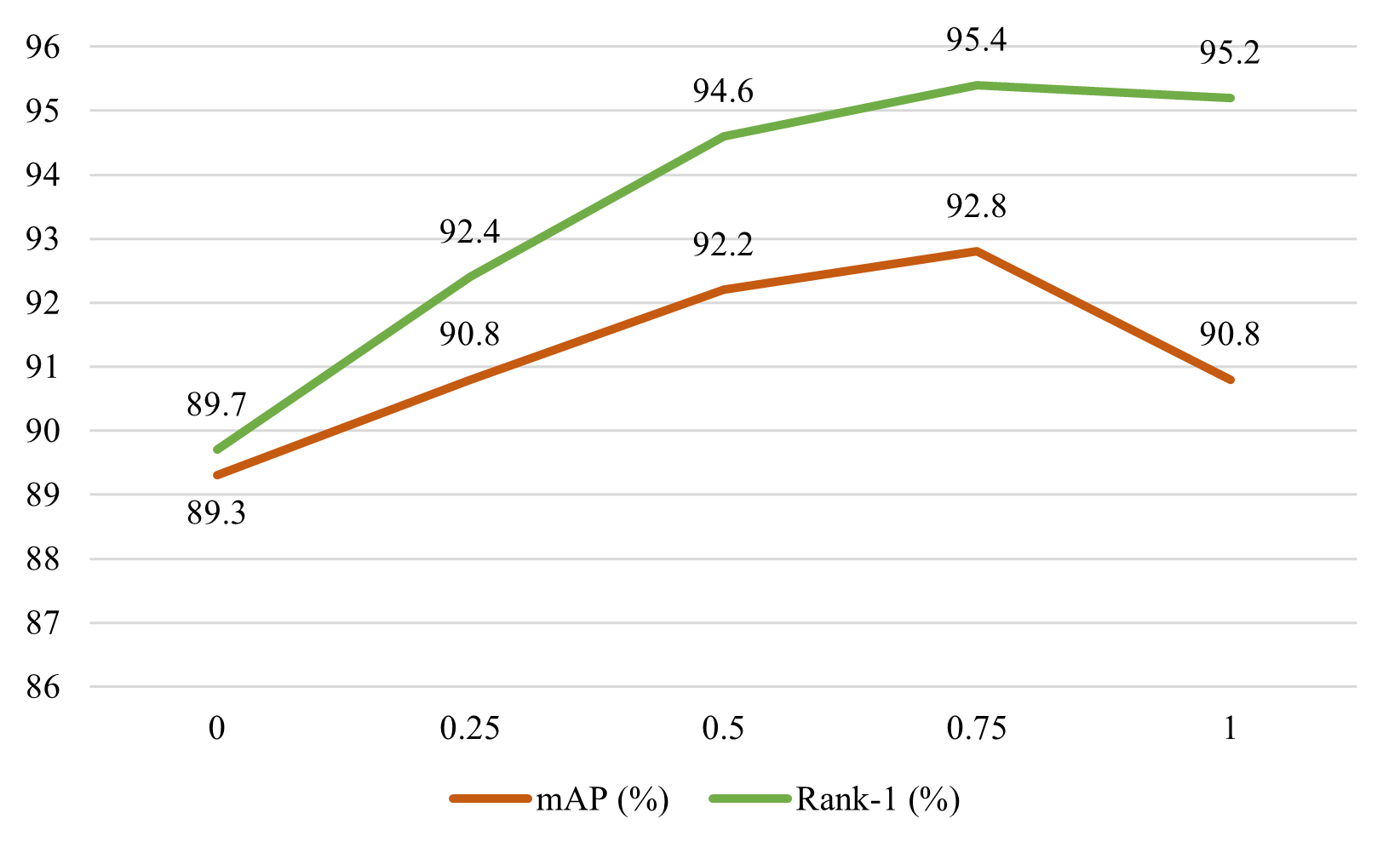}
\caption{Impact of $\gamma$ of DMON on Market-1501 dataset.}
\label{fig:gamma}
\end{figure}

Besides, as shown in Figure~\ref{fig:k}b, we could discover that when $k_2$ is small (e.g., $k_2=1$), the expansion of high-order neighbors is insufficient, adversely affecting the mAP. When $k_2=20$, more sample relationships can be captured, and the performance reaches its peak. As $k_2$ increases further, the accumulation of noise gradually becomes apparent, leading to a decline in performance.

\subsection{Ablation Study}

We conduct an ablation study on the proposed model to validate each contribution for Re-ID tasks. This study analyzes the impact of asymmetric relationship optimization on performance and investigates the optimal order of neighbors in the dynamic multi-order neighborhood.

\begin{table*}[htbp]
\centering
\renewcommand{\arraystretch}{1.4}
\renewcommand\tabcolsep{12pt}
\caption{The Ablation Studies on our proposed modules on Baseline (TransReID).}
\begin{tabular}{l|llllll}

\hline
\multirow{2}{*}{Model} & \multicolumn{2}{l}{Market-1501} & \multicolumn{2}{l}{DukeMTMC} & \multicolumn{2}{l}{MSMT17} \\ \cline{2-7} 
                        & mAP    & Rank-1   & mAP      & Rank-1      & mAP   & Rank-1 \\ \hline
Baseline              & 88.9         & 95.1            & 82.1           & 90.7             & 67.8       & 85.3          \\
+ARO                   & 90.8         & 95.2            & 85.2           & \textbf{91.5}             & 69.1       & 85.9          \\
+DMON                  & 92.3         & \textbf{95.7}            & 86.3           & 90.3             & 73.8       & 84.8          \\
+DMON+ARO              & \textbf{92.8}         & 95.4            & \textbf{87.7}
         & \textbf{91.5}
             & \textbf{75.6}       & \textbf{86.4}          \\ \hline
\end{tabular}
\label{tab:ablation}
\end{table*}

\begin{table}[htbp]
\caption{The Impact of Neighbors with Different Order of DMON Module.}
\begin{center}
\renewcommand\tabcolsep{14pt}
\renewcommand{\arraystretch}{1.2} 
\begin{tabular}{c|cc}

\toprule
N-order neighbors               & mAP & Rank-1 \\
\midrule
1  & \textbf{92.3}     & 95.6        \\
2 & 92.2     & 95.3        \\
3  & \textbf{92.3}     & \textbf{95.7}        \\
4 & 91.9     & 94.8        \\
5  & 91.8     & 94.7        \\
\bottomrule
\end{tabular}
\label{tab:order}
\end{center}
\end{table}

\begin{figure*}[htbp]
\centering
 \includegraphics[width=1\textwidth]{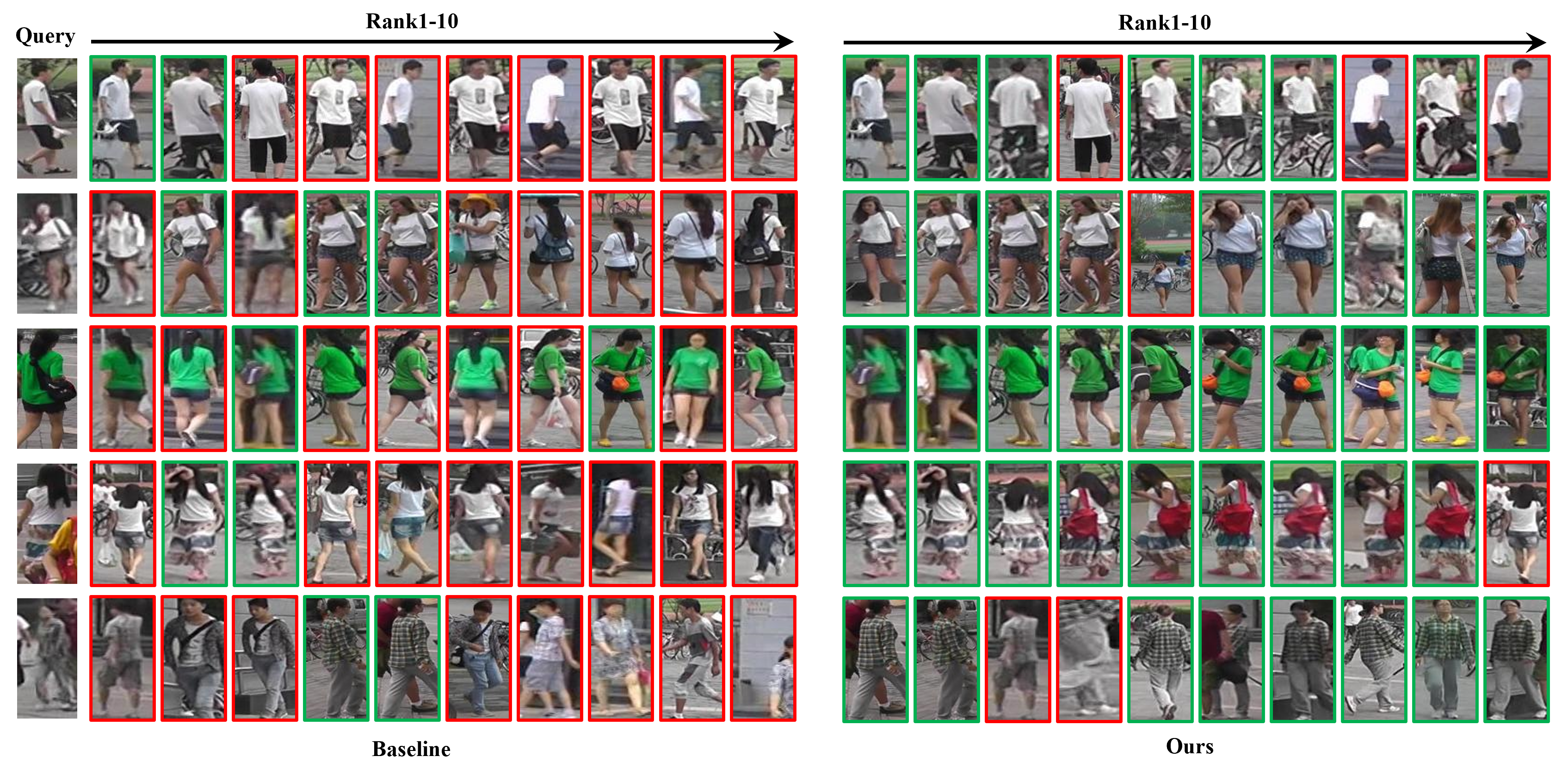}
\caption{Top-10 retrieval visualization of the baseline TransReID and our model on Market-1501.}
\label{fig:rank}
\end{figure*}

\textbf{The Effectiveness of each module.} To evaluate the effectiveness of each component in our model, we observe the results of the ablated models +ARO, +DMON, and the whole model +DMON+ARO compared with the baseline model TransReID. As shown in TABLE~\ref{tab:ablation}, both DMON and ARO modules play significant roles for the whole model from the results of +DMON and +ARO. Besides, DMON shows more performance gains compared to ARO. Furthermore, integrating both DMON and ARO achieves the best performance with mAP improvements of 3.9\%, 5.2\% and 7.5\% on Market-1501, DukeMTMC and MSMT17 compared to the baseline.

\textbf{Optimal Order of Neighbors.} We perform feature enhancement strategy with the neighborhood from first-order to fifth-order neighbors, keeping other hyper-parameters unchanged, and evaluate their influences on Re-ID performance. As shown in Table~\ref{tab:order}, the third-order neighbors demonstrated the best performance improvement on the Market-1501 dataset, confirming the effectiveness of multi-order neighbors in enriching feature representations and enhancing model robustness. However, as the order of neighbors increased further, the marginal contribution to performance gradually diminished.

\subsection{Visualization}
As shown in Figure~\ref{fig:rank}, our model has a significant improvement over the baseline. As can be seen in the figure, for people who are highly confusing, our approach is able to identify the correct person very well, and can even correctly identify a potential target from the unarchived data in the dataset (e.g., the sample marked red by our model in the second row).
\section{Conclusion}
This paper proposes a method for dynamic multi-order neighbor modeling and asymmetric relation optimization, specifically designed to address the complex and challenging feature relationships in pedestrian re-identification tasks. DMON dynamically adjusts the hierarchical relationships among neighbors in the feature space, capturing richer local and global feature dependencies. At the same time, asymmetric relation optimization enhances the expressive power of asymmetric relationships between features through relation weight adjustment and symmetry constraints.

\section*{Acknowledgments}
This work was supported by the National Natural Science Foundation of China (No. 62376016).
{
    \small
    \bibliographystyle{ieeenat_fullname}
    \bibliography{main}
}


\end{document}